\documentclass[journal]{IEEEtran}

\usepackage[noadjust]{cite}
\usepackage{amsmath}
\usepackage{enumerate}
\usepackage{float}
\usepackage{algorithm}
\usepackage{algpseudocode}
\usepackage{lscape}
\usepackage[table]{xcolor}
\usepackage{hyperref}

\usepackage{caption}
\usepackage{subcaption}
\usepackage{graphicx}

\newtheorem{Theorem}{Theorem}
\newtheorem{Lemma}[Theorem]{Lemma}
\newtheorem{Definition}[Theorem]{Definition}

\newtheorem{Remark}[Theorem]{Remark}

\graphicspath{{Figures/}{img/}}

\begin{document}

\title{An efficient projection neural network for $\ell_1$-regularized logistic regression}

\author{Majid Mohammadi, Amir Ahooye Atashin, Damian A. Tamburri
\thanks{Majid Mohammadi and Damian Tamburri are with the Jheronimus Academy of Data Science, Eindhoven University, The Netherlands.
}
\thanks{Amir Ahooye Atashin is with the Department of Computer Engineering, Ferdowsi University of Mashhad, Iran.}}

{}
\maketitle

\begin{abstract}
$\ell_1$ regularization has been used for logistic regression to circumvent the overfitting and use the estimated sparse coefficient for feature selection. However, the challenge of such a regularization is that the $\ell_1$ norm is not differentiable, making the standard algorithms for convex optimization not applicable to this problem. This paper presents a simple projection neural network for $\ell_1$-regularized logistics regression. In contrast to many available solvers in the literature, the proposed neural network does not require any extra auxiliary variable nor any smooth approximation, and its complexity is almost identical to that of the gradient descent for logistic regression without $\ell_1$ regularization, thanks to the projection operator. We also investigate the convergence of the proposed neural network by using the Lyapunov theory and show that it converges to a solution of the problem with any arbitrary initial value. The proposed neural solution significantly outperforms state-of-the-art methods with respect to the execution time and is competitive in terms of accuracy and AUROC. 

\end{abstract}

\begin{IEEEkeywords}
logistic regression, $\ell_1$ regularization, Lyapunov, global convergence. 
\end{IEEEkeywords}

\IEEEpeerreviewmaketitle

\section{Introduction}
Logistic regression is one of the most popular classification techniques with an unconstrained smooth convex problem, which can be solved by the standard algorithms for convex optimization. One of the major problems in classifiers, including logistic regression, is overfitting, which can be mitigated by using a regularization term. $\ell_1$-regularized logistic regression is one of the most popular regularized method, whose minimization problem, for a data set with $n$ samples $\{X,y\}$ where $X \in R^{n\times d}$ and $y \in\{0,1\}^n$, is: 

\begin{align}\label{eq:l1 lr}
    \min_{w} L(X,w) + \lambda \Vert w\Vert_1,
\end{align}
where 
\begin{align}
    &L(X;w) = -\frac{1}{m} \sum_{i=1}^n \bigg(y_i log (\hat{y}_i) + (1-y_i) log(1-\hat{y}_i)\bigg), \cr
    &\hat{y}_i = 1/(1 + exp(-w^Tx_i)),
\end{align}

and $x_i \in R^d$ is the $i^{th}$ sample in the data set. Another option for regularization is $\ell_2$ norm, but $\ell_1$ regularization can be used for feature selection \cite{l1_featureSelection}, and has shown to outperform $\ell_2$ norm when the number of samples is smaller than the number of features \cite{lr_smallsample}. Despite the usefulness of the $\ell_1$ regularization, the drawback of minimization \eqref{eq:l1 lr} is its non-smoothness, placing a serious obstacle to applying the standard algorithms for smooth convex optimization. 

There are several approaches to handle the non-smoothness of problem \eqref{eq:l1 lr}. A popular approach is to use cyclic coordinate descent without computing directly the derivative of minimization \eqref{eq:l1 lr}, which is computationally expensive since computing the objective function of logistic regression entails the costly computation of the sigmoid function \cite{slr_coordinateDescent}. Another approach is to use two auxiliary variables and rewrite $w$ as the difference of two non-negative variables \cite{slr_ip}. This approach, while removing the non-smooth $\ell_1$ norm, increases the dimension of the problem by a factor of two, and subsequently, increases the computational costs. There are also other solvers as well, where the $\ell_1$ norm is replaced by a similar smooth function, and the associated problem is solved using standard techniques for convex programming \cite{slr_smooth}. See also \cite{jmlr_review} and \cite{schimdt_review} for more information on methods and their comparison.

The projection neural networks have shown promising performance in solving optimization problems with intricate constraints \cite{rnn_constrainedProg, rnn_survey}. The neural solution for optimization problems enables us to implement the structure of the network by using VLSI (very large scale integration) and optical technologies \cite{vlsi}, making it applicable to the cases where real-time processing is demanded. Besides, the dynamic system of the recurrent neural networks can be represented by ordinary differential equations (ODEs), so they can be implemented on digital computers as well. An important benefit of neural solutions is their ability to globally converge to an exact solution of the given problem, increasing their suitability for real-world problems. As a result, they have been applied to numerous real-world applications \cite{pnn_svm1, pnn_svm2, pnn_regression1, pnn_regression2, pnn_regression_3, pnn_ip1, pnn_ip2}. Xia and Wang \cite{pnn_svm1} developed a one-layer neural network for support vector classification, which is proved to converge globally exponentially to the optimal solution of the original constrained problem. A simpler neural solution with one-layer structure is developed in \cite{pnn_svm2} and it's global and exponentially convergence is investigated. For regression,  multiple compact neural network models are developed \cite{pnn_regression1,pnn_regression2,pnn_regression_3}, all of which are proved to be globally convergent to a solution. Projection neural networks are applied to other practical problems, such as image restoration \cite{pnn_ip1}, image fusion \cite{pnn_ip2}, robot control \cite{pnn_robot}, non-negative matrix factorization \cite{pnn_nmf}, to name just a few.

A class of projection neural networks is for non-smooth optimization problems, allowing us to use them for solving minimization \eqref{eq:l1 lr} \cite{pnn_nonsmooth1,pnn_nonsmooth2}. Qin et al. \cite{pnn_nonsmooth1} proposed a two-layer projection neural network for non-smooth minimization with equality constraints and showed that the neural solution is globally converge to the optimal solution of the given minimization. A simpler one-layer neural model is developed in \cite{pnn_nonsmooth2} for non-smooth problems with equality and inequality constraints whose global convergence was investigated under mild conditions. While these neural models are applicable to solving problem \eqref{eq:l1 lr}, they are not as efficient as models for smooth optimization problems, and their convergence typically entails some conditions which might not be met in real-world applications. In addition, the non-smoothness in problem \eqref{eq:l1 lr} stems from the $\ell_1$ norm regularization term whose subgradient has a specific form. Therefore, chances are that a more efficient neural model, compared to those for general non-smooth problems, can be developed with guaranteed convergence.    

In this paper, we developed a projection neural network for solving problem \eqref{eq:l1 lr}. Projection neural networks are typically used for constrained problems, where a projection operator is employed to basically make an obtained solution resides inside a feasible region with respect to a set of constraints. While problem \eqref{eq:l1 lr} is unconstrained, this article shows that the proposed neural model converges to the solution of problem \eqref{eq:l1 lr}. The convergence is proved by taking advantage of the Lyapunov theory, where it is demonstrated that the proposed neural solution is stable, and converges to an optimal of problem \eqref{eq:l1 lr} with any arbitrary initial point. The proposed method is computationally simple and its complexity is almost identical to the logistic regression without regularization, allowing us to provide an efficient and scalable implementation for large-scale problems. Extensive experiments on many data sets show that the proposed neural network is significantly fast, and its performance is competitive with state-of-the-art solvers in terms of accuracy and AUROC (area under the receiver operating characteristic curve). The contributions of this paper can be summarized as follows:

\begin{itemize}
    \item The subgradient of the $\ell_1$ norm is adequately dealt with by using a projection operator. As such, a projection neural network is developed, which is shown to converge to a solution to problem \eqref{eq:l1 lr}. In addition, the complexity of the proposed method is similar to the gradient descent for logistic regression without any regularization.

    \item We prove the convergence of the proposed neural method by using the Lyapunov theory for dynamic systems, and will be further shown that the convergence is not reliant on the initialization, in contrast to other methods such as interior-point, where the initialization matters. 

    \item The simplicity of the method allows us to provide an efficient implementation. In particular, our efficient implementation is amenable to parallel computing and execution on the graphics processing unit (GPU), which significantly outperforms other solutions for minimization \eqref{eq:l1 lr}.
    
    \item The Python implementation of the proposed neural network is publicly and freely available\footnote{ \href{https://github.com/Majeed7/L1LR}{https://github.com/Majeed7/L1LR}}.
\end{itemize}

This paper is structured as follows. Section \ref{sec:neural model} presents the proposed neural solution for minimization in \eqref{eq:l1 lr}. Section \ref{sec:stability} presents the stability of the proposed neural solution, while the experiments regarding the proposed neural network on multiple synthetic and real data are discussed in Section \ref{sec:experiments}. Finally, the paper is concluded in Section \ref{sec:conclusion}.

\section{The Neurodynamic Model}\label{sec:neural model}
In this section, we develop a simple projection neural network for problem \eqref{eq:l1 lr}, and it is then extended to take into account a bias term in minimization \eqref{eq:l1 lr}. 

\subsection{Basic Model}
To develop a projection neural network, the Karush–Kuhn–Tucker (KKT) conditions of optimality for problem \eqref{eq:l1 lr} are written as:

\begin{align}\label{eq:kkt}
    \nabla L(X,w) + q = 0, \cr  
\end{align}
where $q = (q_1,q_2,...,q_d)$ are the subgradient of $\lambda \Vert w \Vert_1$ defined as:

\begin{align}\label{eq:subgradient}
    q_i = \partial{(\lambda \Vert w \Vert_1)}_i \left\{\begin{array}{lc}
         =\lambda & w_i > 0 \\
         \in [-\lambda,\lambda] & w_i = 0 \\
         = -\lambda & w_i < 0. 
    \end{array}
    \right.
\end{align}
Using the projection operator, the subgradient in \eqref{eq:subgradient} can be written as \cite{VI_book}:

\begin{align}\label{eq:projection}
    q = P_\Omega(w+q)
\end{align}

where $\Omega = \{z | -\lambda \leq z \leq \lambda \}$, $P_\Omega(z) = [P_\Omega(z_1),P_\Omega(z_2),...,P_\Omega(z_d)]$, and 

\begin{align}
    P_{\Omega}(z_i) = \left\{ 
    \begin{array}{cc}
         \lambda   &z_i > \lambda\\
         z_i  &\vert z_i\vert \leq \lambda \\
         -\lambda  &z_i < -\lambda.
    \end{array}
    \right.
\end{align}

\begin{figure}
    \centering
    \includegraphics[width=.48\textwidth]{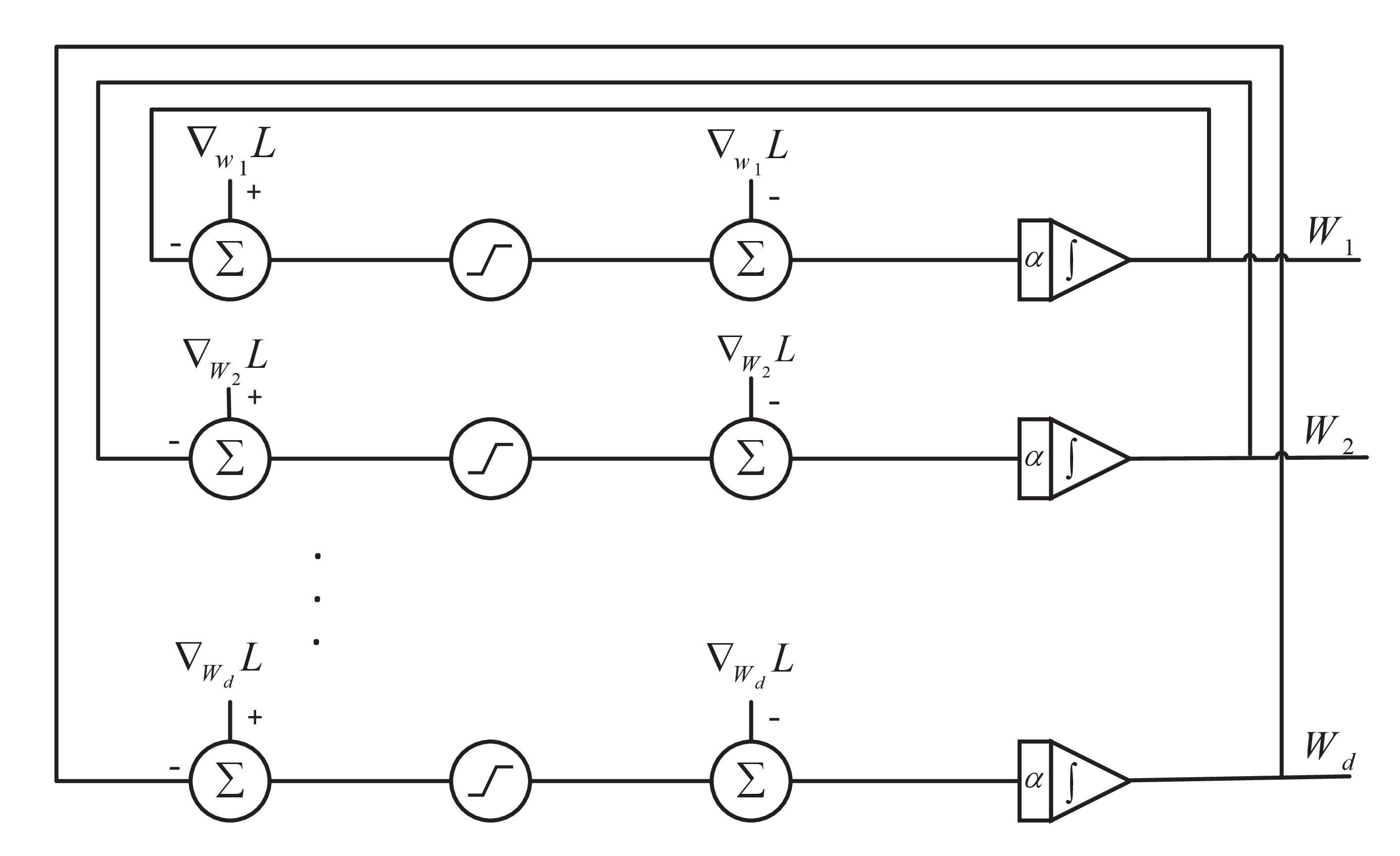}
    \caption{The architecture of the proposed neural solution.}
    \label{fig:architecture}
\end{figure}

By replacing equation \eqref{eq:kkt} in the projection equation in \eqref{eq:projection}, we arrive at the recurrent neural network whose dynamic equation is given by

\begin{align}\label{eq:continuous pgd}
\frac{dw}{dt} &= -\alpha \bigg(\nabla L(X,w) + P_\Omega\big( w - \nabla L(X,w) \big) \bigg)\cr
&= \alpha \bigg(-\nabla L(X,w) + P_\Omega\big( \nabla L(X,w) - w \big) \bigg),
\end{align}

where $\alpha$ is a positive constant and the last equality is derived since $P_\Omega(.)$ is an odd function, i.e., $-P_\Omega(x) = P_\Omega(-x)$. The equation in \eqref{eq:continuous pgd} resembles the projection neural network with a one-layer architecture as shown in Figure \ref{fig:architecture}. According to this figure, we need $2d$ projection (half of which are the sigmoid function for computing $\nabla_{w_i}$), $2d$ summation, $d$ integrators, and some multipliers.

\begin{Remark}
If there is not $\ell_1$ regularization in problem \eqref{eq:l1 lr}, then the following dynamic system representing a recurrent neural network can be used to solve the conventional logistic regression problem:

\begin{align}\label{eq:pgd lr}
    \frac{dw}{dt} &= -\alpha \nabla L(X,w). 
\end{align}
If we discretize this continuous dynamic system, then it will be tantamount to the gradient projection method, which is arguably the most popular optimization algorithm for logistic regression. By juxtaposing equations \eqref{eq:continuous pgd} and \eqref{eq:pgd lr}, it is readily seen that the proposed projection neural network in \eqref{eq:continuous pgd} has the same number of variable as the gradient descent method for logistic regression without regularization, as it is in equation \eqref{eq:pgd lr}. Thus, the proposed neural method does scale up the number of variables even with the $\ell_1$ regularization. The only extra computational overload is $2d$ summations/subtractions for computing the right-hand side in \eqref{eq:continuous pgd}. 
\end{Remark}

\subsection{Bias term computation}
There is typically a bias term $b$ in the standard logistic regression as well, where $\hat{y}$ in problem \eqref{eq:l1 lr} is written as:

\begin{align}
    \hat{y}_i = 1 / \bigg( 1 + exp(-w^Tx_i - b) \bigg). 
\end{align}

To take into account the bias term, one typical is to define $\hat{w} = (w,b)^T$ and $\hat{x}_i = (x_i^T,1)^T$, so the solution to problem \eqref{eq:l1 lr} with $\hat{w}$ and $\hat{x}_i$ would also yield the bias term $b$. However, this problem is slightly different since the absolute value of the bias term also exists in the objective function (which is not sought to be minimized in the standard $\ell_1$-regularized logistic regression). A more proper way is to compute the derivative of the objective function in \eqref{eq:l1 lr} with respect to $b$, which does not include the $\ell_1$ norm. Therefore, the proposed method is turned into:

\begin{equation}\label{eq:pgd with bias}
    \frac{d}{dt}\begin{pmatrix}
        w \\
        b
    \end{pmatrix} = 
    -\alpha \begin{pmatrix}
    
 \nabla_w L(X,w,b) + P_\Omega \big(w - \nabla_w L(X,w,b)\big)  \cr
     \nabla_b L(X,w,b)    \end{pmatrix}
\end{equation}

where $\nabla_w L(X,w,b)$ and $\nabla_b L(X,w,b)$ are the derivatives of the logistic loss function with respect to $w$ and $b$, respectively.
\section{Convergence Analysis}\label{sec:stability}
In this section, the convergence of the proposed projection neural network in \eqref{eq:continuous pgd} is investigated by using the Lyapunov theory. We first provide some necessary definitions and then study the convergence of the dynamic system in \eqref{eq:continuous pgd}.

\begin{Definition}
A vector $w$ is the equilibrium point of the system \eqref{eq:continuous pgd}  if and only if $\nabla L(w,x) = P_\Omega(w - \nabla L(w,x))$. 
\end{Definition}

\begin{Definition}
The dynamic system in \eqref{eq:continuous pgd} is said to be stable at $w^*$ with the initial value $w(0)$ if, for all $\epsilon > 0$, there exists a $\delta(\epsilon) > 0$ such that $\Vert w(k) - w^*\Vert \leq \epsilon$ if $\Vert w(0) - w^* \Vert < \delta(\epsilon)$. 
\end{Definition}

\begin{Lemma}[\cite{VI_book}]\label{lm:p prop}
The projection operator used in equation \eqref{eq:continuous pgd} has the following properties:
\begin{itemize}
\item $(v - P_\Omega(v))^T (P_\Omega(v) - u) \geq 0, \quad \forall v\in R^d, u \in \Omega$.
\item $\Vert P_\Omega(x) - P_\Omega(y)\Vert \leq \Vert x - y \Vert, \quad \forall u,v\in R^d.$
\end{itemize}
\end{Lemma}

We now state the main result for the convergence of the proposed method.

\begin{figure*}[!h]
    \centering
    \includegraphics[width=\textwidth]{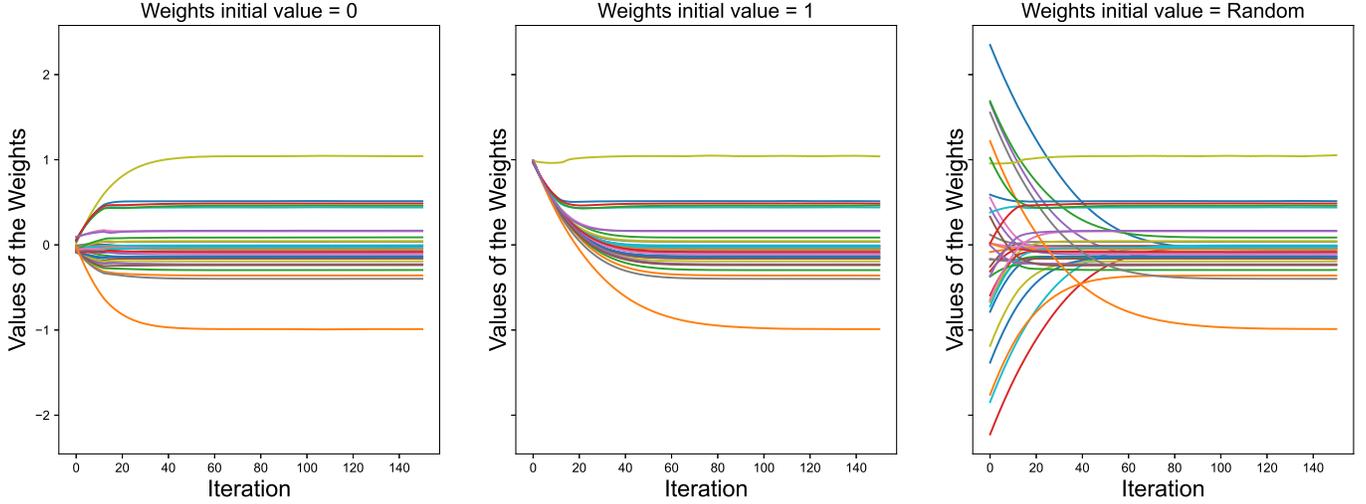}
    \caption{The convergence behavior of the proposed neural network on the \textit{splice} data set with different initial values; From left to right: zero, one, and random initialization. In all cases, $\lambda$ is set to 10.}
    \label{fig:convergence}
\end{figure*}

\begin{Theorem}
For any given initial value $w^0$, the dynamic system in \eqref{eq:continuous pgd} is stable in the sense of Lyapunov and globally converges to equilibrium for any $\alpha > 0$.
\end{Theorem}

\begin{IEEEproof}
We first derived two inequalities that are essential for the proof of the convergence of the dynamic system in \eqref{eq:continuous pgd}. Let $u = \nabla L(X,w^*)$ and $v = \nabla L(X,w) - w$ in the first inequality of Lemma \ref{lm:p prop}, it follows:

\begin{align}\label{eq:conv ineq1}
    \bigg( \nabla L(X,w) - w - P_\Omega\big( \nabla L(X,w) - w \big) \bigg)^T \cr \bigg( P_\Omega\big( \nabla L(X,w) - w \big) - L(X,w^*)   \bigg) \geq 0.
\end{align}

In addition, according to the variational inequality \cite{VI_book}, the solution to the projection equation in \eqref{eq:projection} is tantamount to $w^*$ being held true in the following inequality:

\begin{align}
    &(w-q^*)^T (-w^*) \geq 0, \quad \forall w \in \Omega \quad \cr
    &\Rightarrow  \big(w+\nabla L(X,w^*)\big)^T (-w^*) \geq 0 
\end{align}

Letting $w = -P_\Omega\big(\nabla L(X,w) - w \big) \in \Omega$, the above inequality becomes:

\begin{align}\label{eq:conv ineq2}
    \bigg(P_\Omega\big(\nabla L(X,w) - w \big) - \nabla L(X,w^*) \bigg)^T w^* \geq 0
\end{align}

Summing up the inequalities in \eqref{eq:conv ineq1} and \eqref{eq:conv ineq2}, we arrive at:

\begin{align*}
        &\bigg(P_\Omega\big(\nabla L(X,w) - w \big) - \nabla L(X,w^*) \bigg)^T \cr
        &\qquad \bigg( \nabla L(X,w) - w - P_\Omega\big( \nabla L(X,w) - w \big) + w^* \bigg) \geq 0,
\end{align*}
which results in:

\begin{align}\label{eq:conv ineq last}
    &\bigg( w - w^* + \nabla L(X,w) - \nabla L(X,w^*) \bigg)^T \cr
    & \qquad \bigg( P_\Omega\big(\nabla L(X,w) - w \big) -     \nabla L(X,w) \bigg) \cr
    & \qquad \leq - \big(w-w^*\big) \big( \nabla L(X,w) - \nabla L(X,w^*) \big) - \cr
    & \quad \qquad \Vert P_\Omega\big(\nabla L(X,w) - w \big) - \nabla L(X,w) \Vert^2.
\end{align}

Now consider the following Lyapunov function \cite{lyapunov_function}:

\begin{align}\label{eq:lyapunov}
V(w) &= \int_0^1 \alpha^{-1}(w-w^*)^T \cr
& \quad \bigg(w^* + s(w-w^*) + \nabla L\big(X,w^* + s(w-w^*)\big) \bigg)ds \cr &- \alpha^{-1}(w-w^*)^T\big(w^* + \nabla L(X,w^*)\big) 
\end{align}
where $w^*$ is the equilibrium of \eqref{eq:continuous pgd}. Since $\nabla^2 L(X,w)$ is symmetric and positive semi-definite, $V(w)$ is continuously differentiable and convex in $R^d$. Also, the gradient of $V$ is defined as \cite{lyapunov_function}: 
\begin{align}
    \alpha\nabla V(w) = w - w^* + \nabla L(X,w) + \nabla L(X,w^*).
\end{align}

Based on this Lyapunov function, we have:

\begin{align}\label{eq:stability}
    &\frac{\partial V}{\partial t} = \bigg(\frac{\partial V}{\partial u}\bigg)^T \frac{\partial w}{\partial t} \cr
    & = \bigg( w - w^* + \nabla L(X,w) + \nabla L(X,w^*) \bigg)^T \cr
    & \quad \qquad \bigg( P_\Omega\big(\nabla L(X,w) - w \big) -     \nabla L(X,w) \bigg) \cr 
    & \leq - \big(w-w^*\big) \big( \nabla L(X,w) - \nabla L(X,w^*) \big) - \cr
    &\qquad \Vert P_\Omega\big(\nabla L(X,w) - w \big) - \nabla L(X,w) \Vert^2 \leq 0,
\end{align}
where the last two inequalities are derived based on \eqref{eq:conv ineq last} and the fact that for the convex function $L(X,w)$, we have: 

\begin{align}
    (w - w^*)^T \bigg(\nabla L(X,w) - L(X,w^*) \bigg) \geq 0.
\end{align}

The equation in \eqref{eq:stability} shows that the dynamic system in \eqref{eq:continuous pgd} is stable in the Lyapunov sense. According to the LaSalle's invariant set theorem \cite{LaSalle}, the trajectories of the dynamic system in \eqref{eq:continuous pgd} converge to the largest invariant set $\Phi$ defined as:
\begin{align}
    \Phi = \{w | \partial V(q) / \partial t = 0\}.
\end{align}

To show the global convergence of the proposed method, we need to show $\partial V(w) / \partial t = 0$ if and only if $\partial w / \partial t = 0$. If $\partial w / \partial t = 0$, then

\begin{align*}
    \frac{\partial V}{\partial t} = \bigg(\frac{\partial V}{\partial u}\bigg)^T \frac{\partial w}{\partial t} = 0.
\end{align*}

Conversely, if $\partial V(w) / \partial t = 0$, we have, based on \eqref{eq:stability}, 

\begin{align}
    &\big(w-w^*\big) \big( \nabla L(X,w) - \nabla L(X,w^*) \big) + \cr
    &\quad \Vert P_\Omega\big(\nabla L(X,w) - w \big) - \nabla L(X,w) \Vert^2 = 0.
\end{align}

Since both parts of the above equation are non-negative, it is required that both of the parts become zero, which follows: 

\begin{align}
    P_\Omega\big(\nabla L(X,w) - w \big) - \nabla L(X,w) = 0 \Rightarrow \partial w / \partial t = 0.
\end{align}

Therefore, the dynamic system in \eqref{eq:continuous pgd} converges globally to the solution of problem \eqref{eq:l1 lr}, irrespective of the initial value, and the proof is complete.

\end{IEEEproof}

\begin{figure*}
         \centering
         
         \begin{subfigure}[b]{0.48\textwidth}
           \centering
            \includegraphics[width=\textwidth]{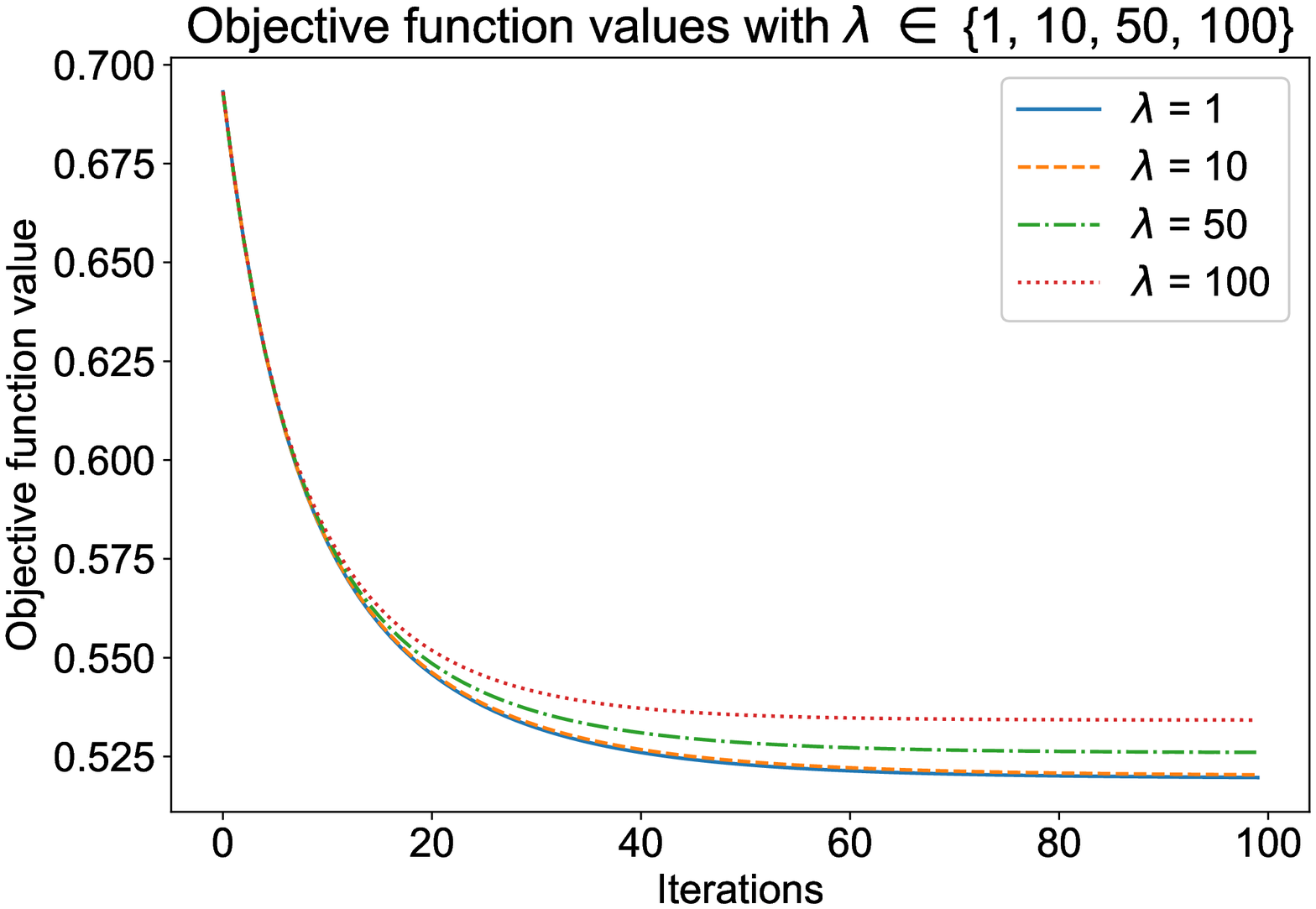}
         \caption{Iterations versus the objective function values.}
         \label{fig:loss}
         \end{subfigure}
         ~ 
         \begin{subfigure}[b]{0.48\textwidth}
            \centering
            \includegraphics[width=\textwidth]{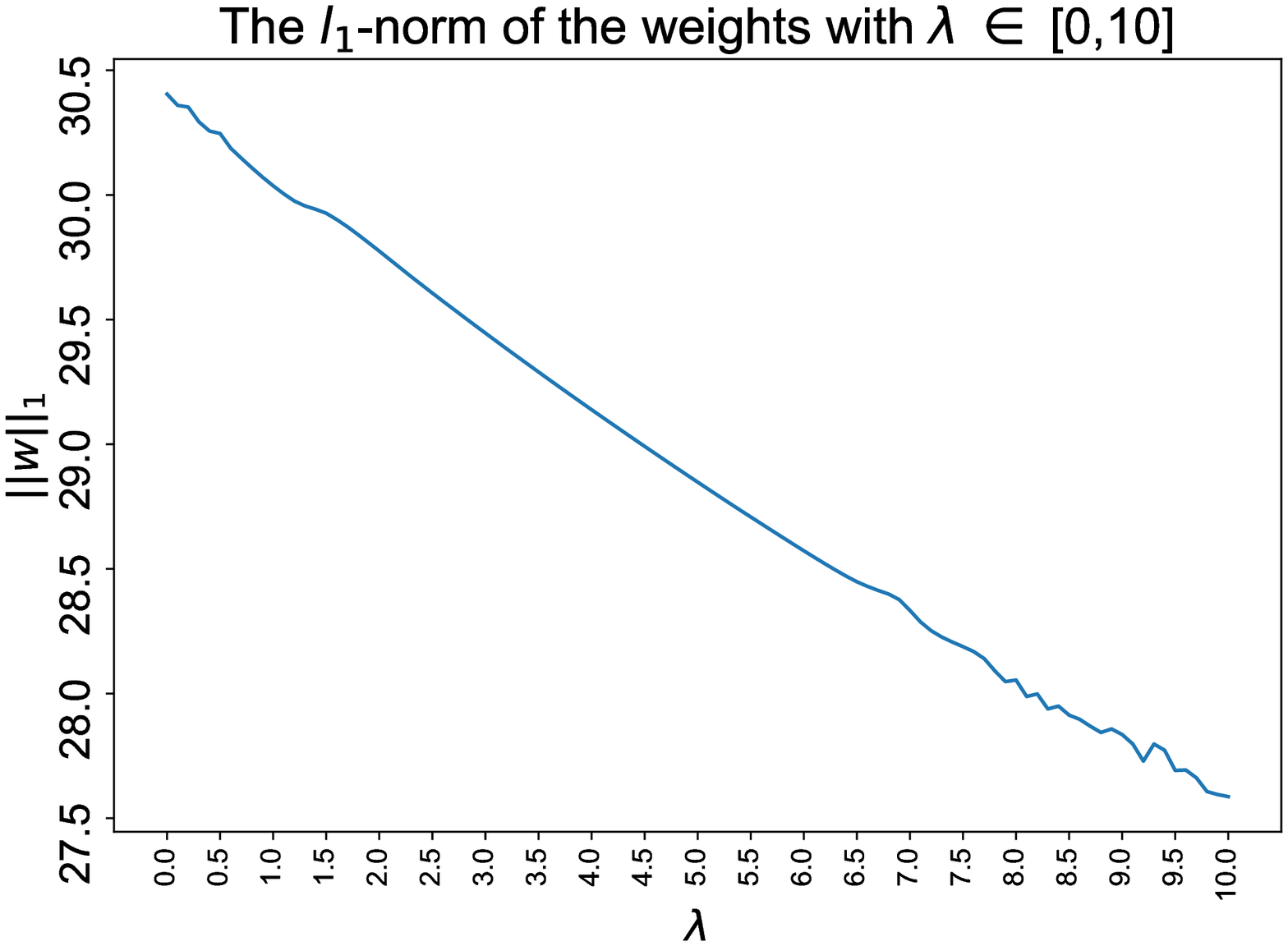}
            \caption{$\lambda$ versus sparsity gauged by $\ell_1$ norm.}
            \label{fig:norm1}
         \end{subfigure}
         
         \caption{The convergence and sparsity behavior of the proposed method on the \textit{madelon} data set. The figures plots: (a) the number of iterations against the objective function value; (b) different values of $\lambda$ against the sparsity of the solution gauged by the $\ell_1$ norm of the coefficient.}
\end{figure*}

\begin{figure*}[!h]
     \centering
     \begin{subfigure}[b]{0.45\textwidth}
         \centering
         \includegraphics[width=\textwidth]{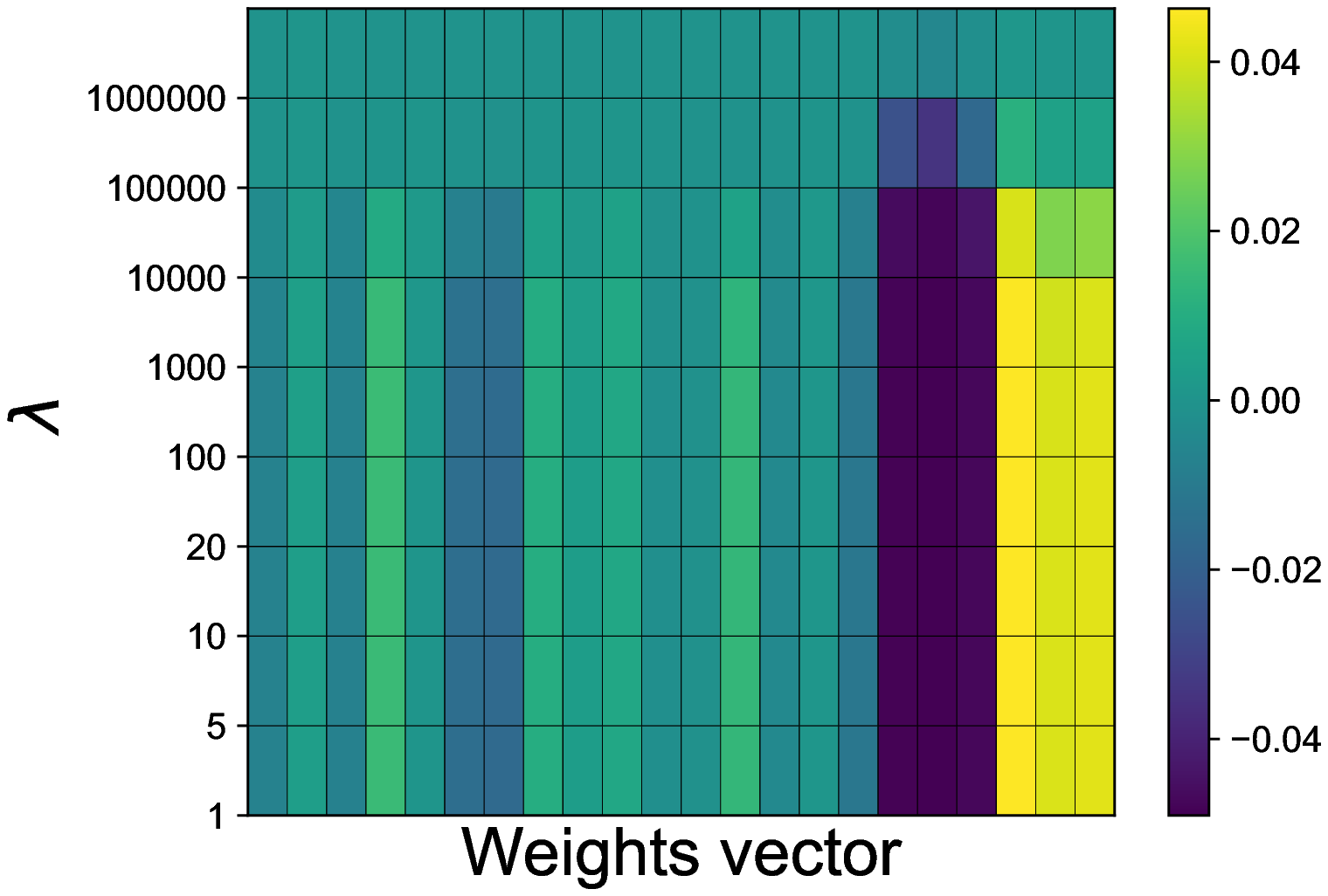}
         \caption{The \textit{ijcnn1} data set. }
         \label{fig:heatmap ijcnn}
     \end{subfigure}
     ~
     \begin{subfigure}[b]{0.45\textwidth}
         \centering
         \includegraphics[width=\textwidth]{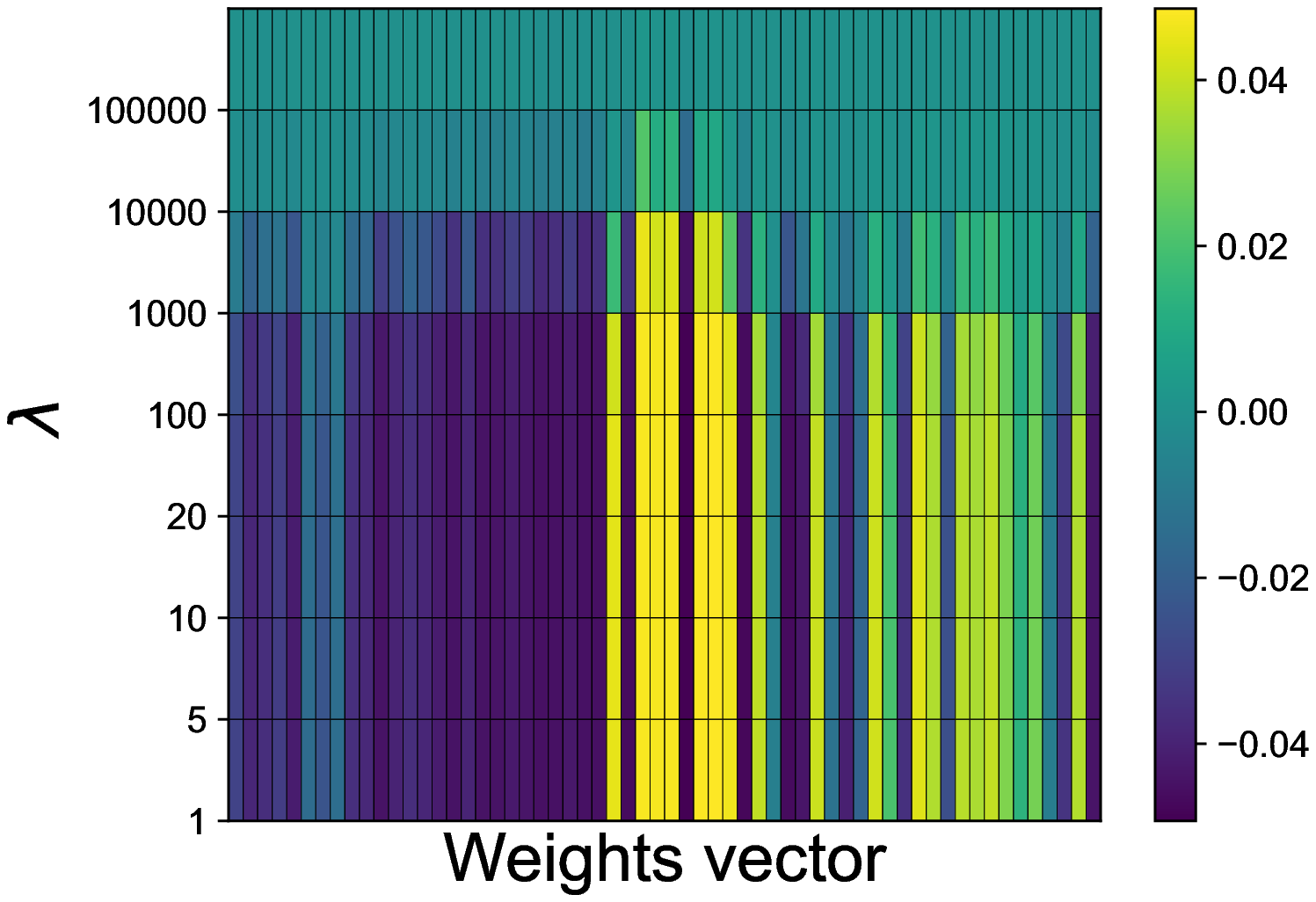}
         \caption{The \textit{splice} data set.}
         \label{fig:heatmap splice}
     \end{subfigure}
     \hfill

        \caption{The heat map of $w$ for different values of $\lambda$ over \textit{ijcnn1} (left) and \textit{splice} (right) data sets.}
        \label{fig:heatmap}
\end{figure*}

\section{Experiments}\label{sec:experiments}
In this section, we first investigate the convergence behavior of the proposed neural network as well as the relationship between the sparsity of solutions and the regularization parameter. We then  compare the performance of the proposed neural solution with several existing $\ell_1$-regularized logistic regression methods in terms of their execution time, accuracy, receiver operating characteristic curve (ROC), and the area under ROC (AUROC) on several data sets.

\subsection{Convergence Analysis}
We first inspect empirically the convergence of the proposed neural network by applying it to the splice data set with different initial values. In particular, we conduct the experiment by using zero, one, and random initialization. Figure \ref{fig:convergence} shows the trajectories of different elements in the weight vector $w$ throughout the optimization procedure. Different elements of $w$ are shown by the same color in the three plots. According to this figure, the trajectories of the proposed neural solution converge to the same solution, regardless of the initial values, which corroborates the global convergence of the proposed network.    

In addition, we investigate the convergence of the proposed method by plotting the objective function value of problem \eqref{eq:l1 lr} in different iterations for various values of $\lambda$, as is shown in Figure \ref{fig:loss}. We also opt for random initialization values to further complicate the convergence of the neural solution.  Figure \ref{fig:loss} illustrates that the objective function value steadily decreases in different iterations and with different values of $\lambda$ until it reaches a minimum. This would also corroborate the global convergence of the proposed neural network given different values for initialization and $\lambda$.

\begin{table}[]
    \centering
    \small
    \caption{The number of features, training, and test samples of the data sets used in the experiments. }
    \label{tab:data sets}
    \resizebox{\columnwidth}{!}{%
    {\rowcolors{2}{white}{gray!50}
    \begin{tabular}{c|c c c}
         \textbf{data set} & \textbf{\# of features} & \textbf{\# of training samples} & \textbf{\# of test samples} \\ \hline
splice           & 60   & 1000& 2175\\
madelon          & 500  & 2000& 600 \\
liver-disorders  & 5    & 145 & 200 \\
ijcnn1           & 22   & 49990 & 91701\\
a1a              & 123  & 1605  & 30956\\
a9a              & 123  & 32561 & 16281\\
leukemia         & 7129 & 38  & 34 \\
gisette          & 5000 & 6000& 1000
\end{tabular}%
}}
\end{table}

\begin{table*}[t]
\center
\caption{The methods' execution time (in seconds) over eight data sets.}
\label{tab:time}
\resizebox{\textwidth}{!}{%
{\rowcolors{2}{white}{gray!50}
\begin{tabular}{c| cccccccc}
\textbf{Method's name} & \textbf{splice} & \textbf{madelon} & \textbf{liver-disorders} & \textbf{ijcnn1} & \textbf{a1a} & \textbf{a9a} & \textbf{leukemia} & \textbf{gisette} \\ \hline
Gauss-Seidel            & 4.13 & 27.67 & 0.05 & 7.79 & 2.90 & 58.42 & 21.90 & 977.58\\
Shooting                & 2.41 & 20.64 & 0.17 & 1.53 & 1.95 & 43.73 & 14.37 & 698.52\\
Gauss-Southwell         & 3.55 & 10.89 & 0.12 & 1.68 & 2.38 & 59.25 & 20.11 & 846.80\\
Grafting                & 1.08 & 20.64 & 0.25 & 2.61 & 2.31 & 50.82 & 21.63 & 649.00\\
SubGradient             & 0.08 & 1.46 & 0.08 & 0.65 & 0.10 & 0.25 & 61.06 & 18.57\\
epsL1                   & 0.04 & 1.07 & 0.06 & 0.36 & 0.18  & 5.68  & 25.80 & 401.36\\
Log-Barrier             & 49.75 & 2.66 & 0.17 & 0.87 & 0.16  & 2.20  & 457.35 & 398.98\\
SmoothL1                & 0.14 & 4.64 & 0.01 & 1.47 & 2.19 & 7.21  & 144.02 & 2214.89\\
SQP                     & 0.06 & 3.40 & 0.19 & 1.20 & 0.62  & 1.06  & 1591.70  & 1585.94\\
ProjectionL1            & 0.04 & 5.20 & 0.04 & 1.51 & 2.51  & 45.25 & 1951.45  & 6483.31\\
InteriorPoint           & 0.22 & 12.34 & 0.09 & 0.53 & 0.42  & 2.67  & 11677.12 & 27362.16 \\
Orthant-Wise            & 0.05 & 20.09 & 0.06 & 1.83 & 3.16  & 46.15 & 143543.25 & 54686.79    \\
Pattern-Search          & 0.16 & 4.33 & 0.21 & 1.34 & 0.57 & 6.03  & 29.52  & 700.61 \\
sklearn                 & 0.01 & 4.63 & 0.01 & 0.53 & 0.62  & 57.32 & 0.04     & 5.94\\
Proposed method         & 0.22 & 0.79 & 0.01 & 0.17 & 0.40  & 1.79  & 0.14     & 1.5 
\end{tabular}
}}%
\end{table*}

\begin{table*}[]
\center
\caption{Accuracy of the methods over eight data sets.}
\label{tab:accuracy}
\resizebox{\textwidth}{!}{%
{\rowcolors{2}{white}{gray!50}
\begin{tabular}{c|cccccccc}
\textbf{Method's name} & \textbf{splice} & \textbf{madelon} & \textbf{liver-disorders} & \textbf{ijcnn1} & \textbf{a1a} & \textbf{a9a} & \textbf{leukemia} & \textbf{gisette} \\ \hline 
Gauss-Seidel            & \textbf{85.47} & \textbf{59.33} & 59.5            & 89.28          & 82.94          & 83.19          & 91.18          & 84.8\\
Shooting                & \textbf{85.47} & 52.50          & 59.0            & 90.50          & 83.45          & 82.46          & 64.71          & 68\\
Gauss-Southwell         & \textbf{85.47} & 58.17          & 59.0            & 90.44          & \textbf{83.90} & 85.01          & 85.29          & 95.2\\
Grafting                & \textbf{85.47} & 57.83          & 59.0            & 91.00          & 83.84          & 84.99          & 82.35          & 94.4\\
SubGradient             & \textbf{85.47} & 56.67          & 59.0            & 9.50           & 24.05          & 23.62          & 58.82          & 50.0\\
epsL1                   & \textbf{85.47} & 56.67          & 59.0           & 91.34          & 83.84          & 84.99          & 76.47          & 98.0 \\
Log-Barrier             & \textbf{85.47} & 56.33          & 59.0           & 91.34          & 83.84          & 84.98          & 85.29          & 97.9 \\
SmoothL1                & \textbf{85.47} & 56.33          & 59.0           & 91.20          & 83.84          & 79.63          & 61.76          & 95.7 \\
SQP                     & \textbf{85.47} & 56.33          & 59.0           & 91.34          & 83.84          & 84.99          & 85.29             & 97.9 \\
ProjectionL1            & \textbf{85.47} & 56.33          & 59.0           & 91.33          & 83.84          & 84.98          & 85.29          & 98.1 \\
InteriorPoint           & \textbf{85.47} & 56.33          & 59.0           & 91.34          & 83.84          & 84.99          & 85.29 	            & 97.9 \\
Orthant-Wise            & \textbf{85.47} & 56.33          & 59.0           & 91.33          & 83.84          & 84.97          & 85.29 	     &\textbf{98.3} \\
Pattern-Search          & \textbf{85.47} & 56.33          & 59.0            & 91.34          & 83.84          & 84.99          & 88.24      & 98.2 \\
sklearn                 & \textbf{85.47} & 56.67          & 59.00           & 88.23         & 83.83          & 84.99          & 88.23      &97.9\\
Proposed method         & \textbf{85.47} & 59.0          & \textbf{61.50}  & \textbf{91.90} & 83.62          & \textbf{85.03} & \textbf{94.11}       & 98.0 \\
\end{tabular}
}}%
\end{table*}
\subsection{Sparsity}
We now look into the sparsity of the solutions given by the proposed method by subjecting the \textit{madelon} data set to the proposed method. Figure \ref{fig:norm1} plots the value of $\lambda$ against the $\ell_1$ norm of the solution. It is readily seen from this figure that the sparsity increases as expected when the value of $\lambda$ is set to a higher value.  This figure supports the sparsity induced by the proposed method, which is controlled by the parameter $\lambda$.

We also plot the coefficient vector $w$ for different values of $\lambda$ by using a heat map. Figure \ref{fig:heatmap} shows the heat map of $w$ for different values of $\lambda$ over \textit{ijcnn1} and \textit{splice} data sets. The heat map shows that the vast majority of elements in $w$ take on zero or infinitesimal values, shown by greenish color, while there are some few elements with non-zero and bigger values, shown as yellowish colors. As expected, if the values of $\lambda$ exceed a threshold, then the resulting $w$ will become all zero, as shown at the top row of the two plots in Figure \ref{fig:heatmap}. 

\begin{figure*}[!h]
    \centering
    \includegraphics[width=.9\textwidth]{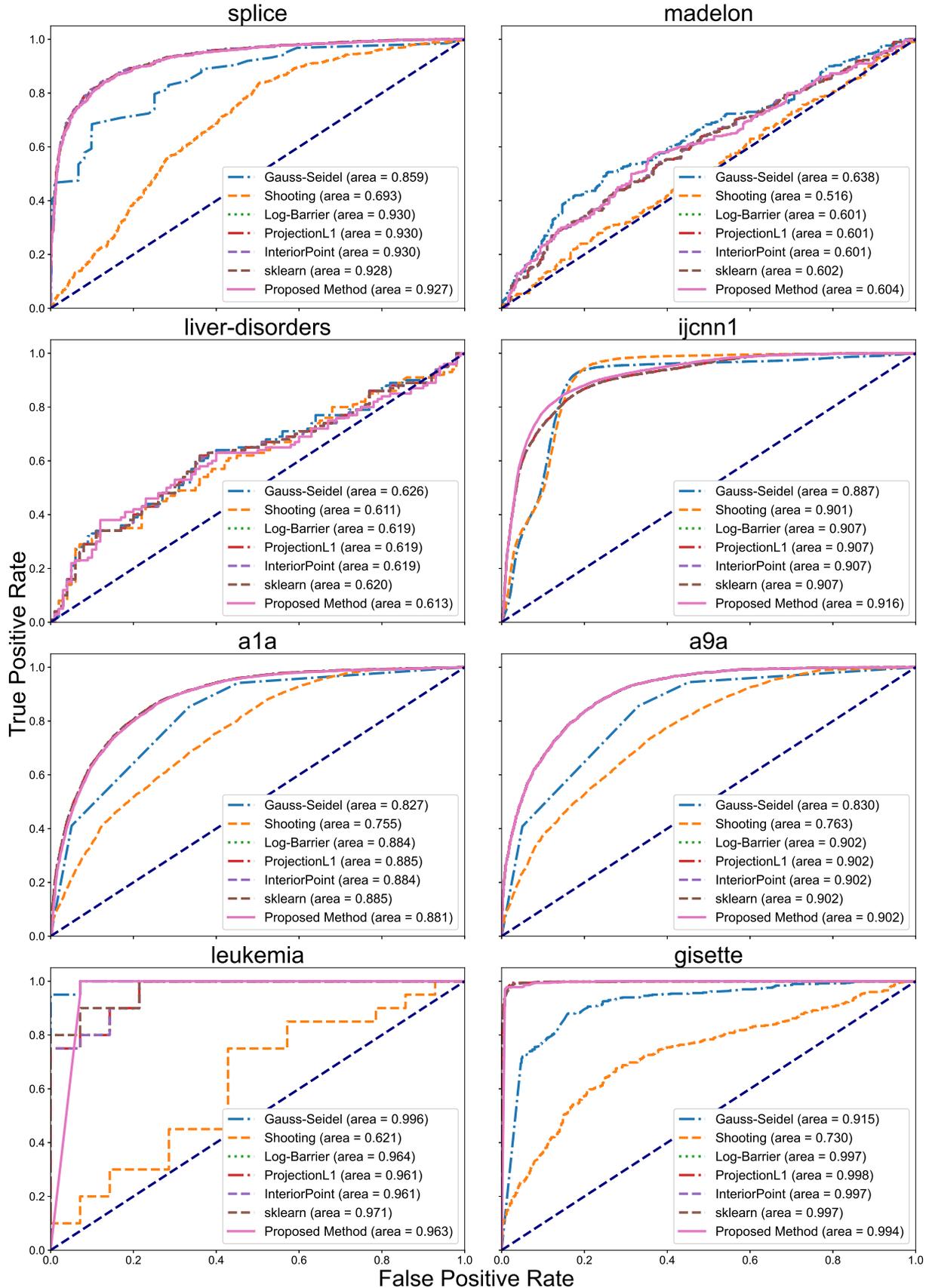}
    \caption{The ROC curve and AUROC of the proposed neural network and six other methods on six data sets. The x-axis of each plot represents the false positive rate, while the y-axis is the true positive rate. }
    \label{fig:roc}
\end{figure*}

\subsection{Comparison of real data sets}
We now compare the performance of the proposed neural solution with other state-of-the-art solvers on several real data sets. The comparison is made in terms of execution time, classifier accuracy, ROC, as well as AUROC. We first introduce the methods and data sets being used. 

\paragraph*{Methods} For the comparison study, we select several methods, each with a different approach to handle the $\ell_1$ norm. The methods are Gauss-Seidel \cite{gauss_seidel}, Shooting \cite{shooting}, Gauss-Southwell \cite{gauss_southwell}, Grafting \cite{grafting}, Subgradient \cite{schimdt_review}, epsL1 \cite{sqp}, Log-Barrier \cite{schimdt_review}, SmoothL1 \cite{slr_smooth}, SQP \cite{sqp}, ProjectionL1 \cite{schimdt_review}, InteriorPoint \cite{slr_ip}, Orthant-wise \cite{schimdt_review}, and Pattern-Search \cite{schimdt_review}, all implemented in MATLAB and are freely available \cite{l1general}. We also use the implementation for logistic regression in \textit{scikit-learn}, and refer to it as \textit{sklearn} \cite{scikitlearn}. The core of \textit{scikit-learn} methods have been implemented in $C/C\!+\!+$, and the implementation of logistic regression takes advantage of an inherent feature and subset selection, which makes it significantly faster.

\paragraph*{Data sets and experimental setup} We subject the proposed method and other methods to eight data sets: \textit{leukemia}, \textit{live-disorders}, \textit{madelon}, \textit{splice}, \textit{gisette}, \textit{ijcnn1}, \textit{a1a}, and \textit{a9a}. The data sets have their own training and test partitions, which are used for the training and testing of all the methods. The information regarding these data sets is tabulated in Table \ref{tab:data sets}.

\paragraph*{Execution time comparison} We first compare the methods in terms of the time they consumed to solve problem \eqref{eq:l1 lr}. For doing so, we set the stopping criterion of all the methods to $10^{-6}$. Table \ref{tab:time} shows the execution time of the methods to solve problem \eqref{eq:l1 lr} for different data sets. For those of data sets with a limited number of features and training samples, all the methods behave efficiently and converge to an optimal in a relatively fast manner, e.g., the performance of methods over the data sets \textit{splice}, \textit{madelon}, \textit{liver-disorders}, and \textit{a1a}. Even for a data set with a large number of the training samples like \textit{ijcnn1}, the performance of the solvers in terms of their execution time is quite competitive, mainly because the dimensions of the solvers are typically linear in the number of features, which is $22$ for the $ijcnn1$ data set. But when the number of features and training samples increases together, many of the solvers fail to produce a solution in a reasonable time frame. For example, it takes more than three and seven hours for the \textit{InteriorPoint} method to generate a result on the \textit{leukemia} and \textit{gisette} data sets, respectively. Or it takes even more for the \textit{Orthant-Wise} method to generate results for the same data sets. Among other methods, \textit{SubGradient} has a very acceptable performance in terms of execution time in generating acceptable results, but we also need to investigate how good its result is regarding accuracy and AUROC. The proposed neural solution, on the other hand, shows a significantly better performance compared to other methods in terms of execution time and demonstrates a fast performance regardless of the size of the data set. In particular, it is especially competitive with \textit{sklearn} despite its efficient implementation (which includes feature and subset selection in $C/C\!+\!+$). A surprising point is also on the execution time of \textit{sklearn} for the \textit{a9a} data set, which shows that a significant increase in both feature and training samples number would adversely affect the performance of this solver. The proposed method demonstrates a stable performance on different data sets with a different number of features and/or training samples.     

\paragraph*{Accuracy comparison} While the execution time of the proposed method is quite superior to other solvers, it is also essential to compare its performance in terms of accuracy. Table \ref{tab:accuracy} shows the accuracy of different methods on different data sets\footnote{Some of the numbers are the same in this table, simply because we round the number to two decimal digits and that causes us to ignore the infinitesimal differences between the accuracy of two solvers.}. According to this table, the accuracy of the proposed method is better than or equivalent to other solvers on five data sets: \textit{splice}, \textit{liver-disorders}, \textit{ijcnn1}, \textit{a9a}, and \textit{leukemia}. For the remaining three data sets, the accuracy of the proposed method is quite competitive with the best-performing solvers. In particular, the accuracy difference between the proposed neural solution and best-performing solvers for the three data sets (i.e., \textit{Gauss-Seidel} for \textit{madelon}, \textit{Gauss-Southwell} for \textit{a1a}, and \textit{Orthant-Wise} for \textit{gisette}) is less than $1\%$, making the performance of the proposed neural solution very competitive with the state-of-the-art solvers in terms of accuracy. At the same, these methods require plenty of time to generate a result. For example, \textit{Orthant-wise} takes more than 15 hours to generate results on the \textit{gisette} data set, while its accuracy is only $0.3\%$ higher than that of the proposed neural solution.

\paragraph*{ROC curve and AUROC comparison} We now compare the proposed neural solution in terms of ROC curve and AUROC. Figure \ref{fig:roc} plots the ROC curves of the methods over eight data sets, the legend of which shows the AUROC of each method as well. As in the ROC curve, a deviation from diagonal shows a better performance, this figure also supports that the performance of the proposed neural solution is competitive with other solvers. In particular, the proposed neural network is the best-performing method on \textit{a9a} and \textit{ijcnn1}. In addition, its difference with the top-performing method on \textit{a1a}, \textit{splice}, and \textit{gisette} data sets are very infinitesimal with a difference in AUROC less than $0.004$. However, Gauss-Seidel has shown better performance on the \textit{madelon} data set with a higher margin. This experiment also upholds the reasonable performance of the proposed neural network compared to the state-of-the-art solvers. Therefore, the conclusion can be drawn that the proposed neural solution can provide a reliable solution to the logistic regression with $\ell_1$ regularization, while it is significantly fast and is scalable for large-scale problems. 

\section{Conclusion}\label{sec:conclusion}
This paper presented a simple yet efficient projection neural network to solve $\ell_1$-regularized logistic regression. The proposed solver utilizes the projection operator, which is typically used in constrained optimization, for the unconstrained but non-smooth problem. The global convergence of the proposed method is guaranteed by using the Lyapunov theory, and its competitive performance with other state-of-the-art solvers was demonstrated on several real data sets. 

The proposed neural solution was developed without using any auxiliary variable, so the dimension of the solver is the same as the dimension of the original problem. This means that the complexity of the proposed neural network for $\ell_1$-regularized logistic regression is the same as that without regularization, making the resulting projection neural network very efficient in terms of execution time and memory consumption.  

For the future, similar projection neural solutions can be tailored for other non-smooth problems, such as Lasso, where only the logistic loss function is replaced by the least square. Another extension of this method can be used to solve the support vector machine problem in primal, by incorporating the subgradient of the maximum function by a projection operator.

Another area of interest is the training of the multi-layer feedforward neural networks, where $\ell_1$ regularization is of the highest interest, especially in preventing overfitting. The result of this research can be further used, probably with some modification, in order to develop more efficient training algorithms for feedforward neural networks when they regularize the weights with an $\ell_1$ penalty.

Methods for tuning the learning rate $\alpha$ in the projection neural solution should be considered, since a proper way of tuning this parameter significantly affects the convergence behavior. One solution is to use model-agnostic meta-learning, which has been successfully applied to various learning models.

\bibliographystyle{IEEEtran}
\bibliography{ref}
\end{document}